\documentclass[conference]{IEEEtran}
\IEEEoverridecommandlockouts

\usepackage{cite}
\usepackage{amsmath,amssymb,amsfonts}
\usepackage{algorithmic}
\usepackage{graphicx}
\usepackage{textcomp}
\usepackage{xcolor}
\usepackage{booktabs}  
\usepackage{array}    
\usepackage{caption}

\usepackage{amsmath,amssymb,amsfonts}
\usepackage{algorithmic}

\usepackage{hyperref}
\usepackage{colortbl} 
\usepackage{multirow}
\usepackage{placeins}  
\usepackage{subcaption} 
\usepackage{enumitem} 
\usepackage{float}

\def\BibTeX{{\rm B\kern-.05em{\sc i\kern-.025em b}\kern-.08em
    T\kern-.1667em\lower.7ex\hbox{E}\kern-.125emX}}

\title{FedLoRA-Optimizer: Federated LoRA Fine-Tuning with Global and Local Optimization in Heterogeneous Data Scenarios}

\begin{document}

\author{
    \textbf{Jianzhe Zhao}\(^1\) \quad 
    \textbf{Hailin Zhu}\(^1\) \quad 
    \textbf{Yu Zhang}\(^1\) \quad 
    \textbf{Ziqi Chen}\(^1\) \quad 
    \textbf{Guibing Guo}\(^1\) \\  
    \vspace{0.5em}  
    \(^1\) Northeastern University
}

\maketitle

\begin{abstract}
Federated efficient fine-tuning has emerged as an approach that leverages distributed data and computational resources across nodes to address the challenges of large-scale fine-tuning and privacy preservation. The Low-Rank Adaptation enables efficient fine-tuning of large-scale pre-trained models by introducing trainable low-rank matrices into weight updates.However, in heterogeneous data scenarios, client drift weakens the generalization of the global model, and local models often fail to meet the personalized needs of individual clients.Moreover, existing federated LoRA efficient fine-tuning techniques overlook fine-grained analysis of the tuning matrices. To address this, we conducted preliminary experiments and found that different LoRA matrices exhibit different sensitivity to changes in the direction and magnitude of their vectors.We thus propose a fine-grained federated LoRA tuning method. By fine-tuning the more sensitive directional vectors in the A matrix, which encode shared knowledge, our method learns shared features more effectively across clients and enhances global generalization. Simultaneously, by fine-tuning the more sensitive magnitude vectors in the B matrix, which encode personalized knowledge, our method better captures personalized knowledge, enabling detailed adaptation to local data. The method uses a pipeline combining global and local optimizers. Global optimization further improves local models, achieving collaborative optimization between global and local levels. This improves both the generalization ability of the global model and the personalized adaptation of local models under heterogeneous data scenarios. Experiments on Databricks-Dolly-15k and Natural Instructions with LLaMA2-7B and Deepseek-7B confirm that our method improves global performance by 0.39\% and local performance by 0.59\%.
\end{abstract}


\section{Introduction}

Federated learning is a distributed machine-learning framework designed to overcome data-silo issues by enabling multiple parties to collaboratively train and optimize a global model without sharing their local data. With the rapid advancement of artificial intelligence, large-scale pre-trained models have garnered widespread attention in academia and industry due to their outstanding performance. However, training such models from scratch often faces severe computational bottlenecks, given their enormous parameter counts~\cite{touvron2023llama}. In this context, efficient fine-tuning techniques have emerged as a key approach to enhancing the practical usability of large models while reducing resource consumption. The core idea behind these methods is to freeze the backbone model’s parameters and introduce only a small set of task-specific weights, thereby significantly improving efficiency during the fine-tuning phase and subsequent deployment. This paradigm has driven the broad adoption of large language models in real-world applications. Among these approaches, low-rank adaptation (LoRA) has attracted particular attention because of its parameter efficiency, training stability, and strong adaptability. It has inspired a host of derivative methods that demonstrate excellent empirical performance~\cite{hu2022lora,liu2024dora,tian2024hydralora}.

Combining a cloud-to-edge federated learning architecture with efficient fine-tuning strategies for large models has drawn significant interest due to its resource utilization and privacy protection advantages. Prior studies have incorporated low-rank adaptation techniques such as LoRA into federated environments and achieved satisfactory fine-tuning results. However, in heterogeneous data scenarios, the global model’s generalization ability often remains limited, and each client’s personalized model may underperform, leading to an overall degradation in fine-tuning effectiveness~\cite{sun2024improving}.

Furthermore, most existing federated LoRA fine-tuning methods focus primarily on model architecture design, with limited attention given to fine-grained analysis of the changes in the fine-tuning matrices. To better understand the magnitude and directional variations of the matrices during the fine-tuning process, we designed and conducted a series of exploratory experiments. In a federated learning environment with heterogeneous data, we applied the LoRA fine-tuning method to various downstream and global tasks, systematically recording the two fine-tuning matrices' magnitude and directional vector changes. The experimental results show that the directional vector changes of matrix A are approximately 1.7 times greater than matrix B's. In comparison, the magnitude changes of matrix B are about 41 times greater than matrix A's.

Based on this finding, we propose a new federated LoRA framework, FedLoRA-Optimizer, which dynamically adjusts the magnitude and direction vectors of the fine-tuning matrices during training, with the aim of achieving a more generalizable global model and a more performant personalized model. We empirically evaluated LLaMA2-7B and Deepseek-7B models using the Databricks-Dolly-15k and Natural Instructions datasets. The results demonstrate that under identical task settings, FedLoRA-Optimizer achieves consistent accuracy improvements compared to traditional LoRA methods. In summary, our main contributions are as follows:
\begin{itemize}
\item We conduct fine-grained empirical analyses of the directional and magnitude variations of LoRA fine-tuning matrices. Experiments on different downstream tasks show that the directional variation in matrix A is approximately 1.7 times larger than in matrix B. In contrast, the magnitude variation in matrix B is about 41 times greater than in matrix A.

\item We propose a fine-grained federated fine-tuning method tailored to heterogeneous data scenarios. By separately optimizing the highly sensitive directional vectors in matrix A and the highly sensitive magnitude vectors in matrix B across different tasks, our method significantly enhances the generalization of the global model and the adaptability of local models.

\item We validate the rationale of our method through parameter change analysis and ablation studies. Subsequently, formal experiments on LLaMA2-7B and Deepseek-7B using the Databricks-Dolly-15k and Natural Instructions datasets show that our method improves accuracy by approximately 0.39\% on global tasks and 0.59\% on local tasks compared to traditional LoRA approaches.
\end{itemize}

\section{Related Works}
\textbf{Preliminary}: In the field of parameter-efficient fine-tuning, Liu et al. identified key limitations of LoRA: first, its update pattern is overly simplistic—magnitude and direction are positively correlated—failing to capture the negative correlation observed in full fine-tuning, which reduces its adaptability to downstream tasks; second, it does not effectively separate the magnitude and direction of pre-trained weights, hindering fine-grained modeling for heterogeneous tasks. To address these issues, DoRA introduces a Direction-Magnitude (D-M) decomposition strategy, which factorizes the pre-trained weights into separate magnitude and direction components, as shown in Equation (1).
\begin{equation}
    W = m \frac{V}{\lVert V \rVert_c}
\end{equation}

Here, $m$ is the magnitude vector, $V$ is the directional matrix, and the expression $\lVert \cdot \rVert_c$ denotes the vector-wise norm of a matrix across each column.

\begin{figure*}[!htbp]
  \centering
  \includegraphics[width=\textwidth]{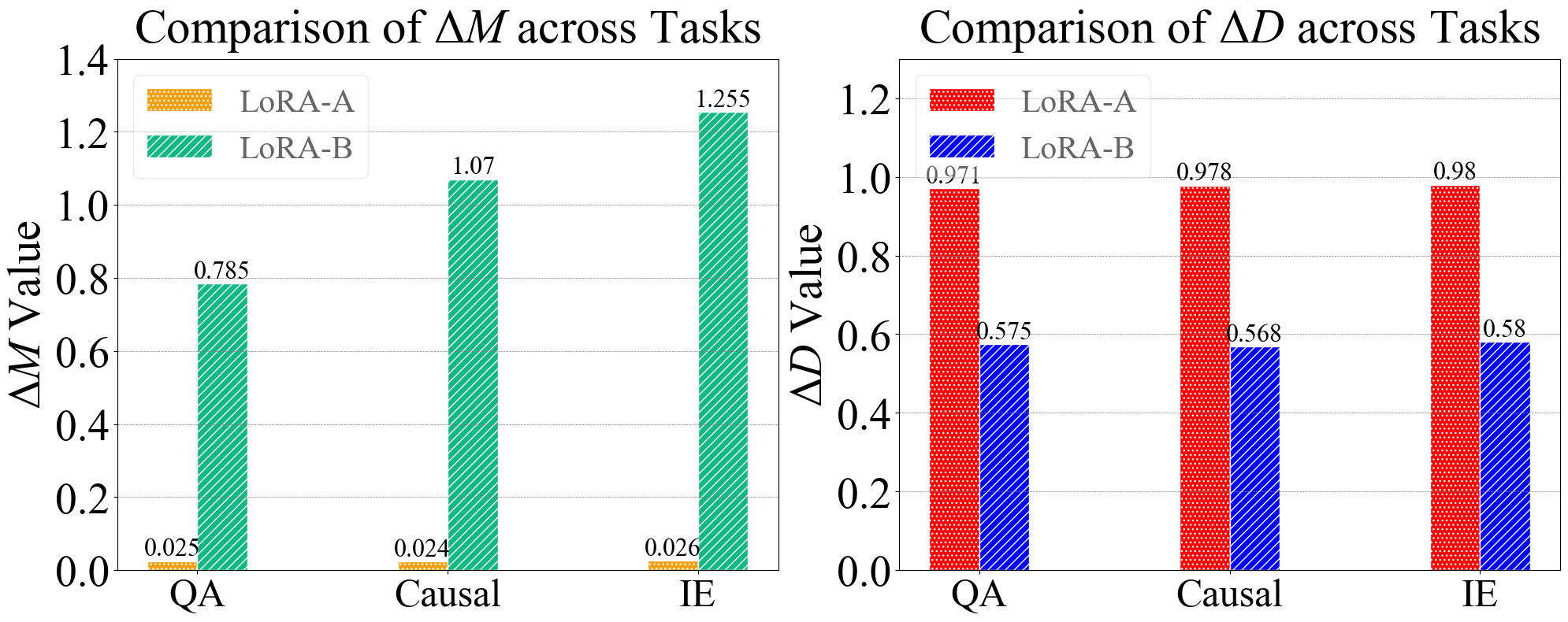}
  \caption{Sensitivity of federated LoRA fine-tuning to magnitude changes and direction changes in A and B matrices; we conducted training using the Databricks-Dolly-15K dataset on the LLaMA2-7B model.}
  \label{fig:your_label1}
\end{figure*}

\textbf{Federated Learning}: In federated learning, global optimization aggregates local client updates into a global model. FedAvg optimizes this by averaging updates but suffers under data heterogeneity~\cite{mcmahan2017communication}. To address heterogeneity, methods like data augmentation, adaptive optimization, and meta‐learning have been proposed~\cite{hao2021towards,yeganeh2020inverse,li2021fedsae,li2020federated,li2020communication,li2022federated,corinzia2019variational,chen2020federated}. FedProx adds a proximal term to constrain each local update’s deviation from the global model, improving convergence and accuracy on non‐IID data~\cite{li2020federated}. SCAFFOLD employs control variates to correct “client drift,” achieving robust convergence under heterogeneity with fewer communication rounds~\cite{karimireddy2020scaffold}. Other approaches reweight client updates to mitigate uneven distributions~\cite{kou2024pfedlvm}; data‐sharing introduces small auxiliary datasets to balance information across participants~\cite{he2022sparseadapter}; and personalized federated learning builds bespoke client models to capture unique local characteristics, boosting predictive performance~\cite{goetz2020federated}. Although effective under heterogeneity, these methods are not specifically tailored for federated fine‐tuning.

\textbf{Large Model Fine-Tuning}: Full-parameter fine-tuning of large models incurs high computational and storage costs, so parameter‐efficient methods are crucial. Houlsby et al. proposed Adapters, small trainable modules inserted into each Transformer layer, to enhance performance with minimal tuning~\cite{lester2021power}. Hu et al.’s LoRA decomposes weight matrices into low-rank components, updating only bypass modules to drastically reduce tunable parameters~\cite{hu2022lora}. Liu et al. further decompose LoRA matrices into ``magnitude + direction" for finer, efficient tuning~\cite{liu2024dora}. Tian et al. observed that LoRA’s A matrix captures global common knowledge.

The B matrix is more suitable for expressing personalized features, leading to a new LoRA fine-tuning architecture~\cite{tian2024hydralora}. Prompt-Tuning guides the model toward specific tasks through carefully designed textual prompts, improving NLP performance~\cite{lester2021power}. SparseAdapter applies extensive parameter pruning at initialization, leveraging sparsity-inducing techniques to maintain performance while significantly lowering computational cost, and augments capacity by expanding bottleneck dimensions through higher sparsity ratios~\cite{he2022sparseadapter}. Liu et al. proposed p-tuning, which concatenates trainable continuous prompt embeddings with discrete prompts as input to a pretrained model; these embeddings are adjusted via a prompt encoder (e.g., an LSTM or MLP) through backpropagation to minimize task-specific loss~\cite{liu2024gpt}. Building on this, p-tuning v2 injects continuous embeddings as prefix tokens at every layer, narrowing the performance gap with full fine-tuning, especially improving results for smaller models and more challenging tasks~\cite{liu2021p}. Although these methods are effective for large-model fine-tuning, performance may still degrade substantially when downstream tasks diverge greatly from the pretraining data.

\textbf{Parameter-Efficient Federated Fine-Tuning}: Researchers have proposed parameter-efficient federated fine-tuning methods to jointly train large models on distributed data while reducing communication and preserving privacy. Qin et al. combine zeroth-order optimization with shared random seeds to enable clients to fine-tune billion-parameter models by transmitting only kilobytes of updates~\cite{qin2023federated}. Wang et al. show that, for foundation models exceeding one billion parameters, a single round of communication can match multi-round performance~\cite{lester2021power}. Sun et al. note that naive LoRA in federated settings leads to unstable convergence~\cite{sun2024improving} and propose FFA-LoRA, which fixes one low-rank matrix while fine-tuning the other to improve stability and further lower communication. Nonetheless, data heterogeneity, where client datasets vary widely in quality and distribution, continues to hinder global generalization, leading to slower convergence and reduced accuracy~\cite{feng2024adapter,tian2022harmony}.

\section{Motivation and Observation}

To develop a federated fine-tuning approach tailored for heterogeneous data environments, we focus on analyzing the directional and magnitude variations of the matrices involved in the LoRA fine-tuning process across different downstream tasks. To this end, we design and conduct experiments to uncover the underlying mechanisms of LoRA adaptation. Specifically, we conduct experiments on the LLaMA2-7B model, fine-tuned using the LoRA method on three representative downstream tasks: Causal task, IE task, and QA task, which are selected from the Databricks-Dolly-15k dataset. We also aggregate these tasks to form a global task setting, from which the corresponding LoRA adapter matrices A and B are obtained.
Subsequently, inspired by the decomposition approach proposed by Liu et al.~\cite{liu2024dora}, we compute the changes in magnitude vector and direction vector of the A and B matrices for each downstream task relative to the global task. The computation is written as follows:

\begin{equation}
    \Delta M_{A 11}^{\mathrm{t}}=\frac{\sum_{n=1}^{k}\left|m_{A l l}^{n, t}-m_{0}^{n}\right|}{k}
\end{equation}

\begin{equation}
    \Delta D_{A 11}^{t}=1-\cos \left(V_{A l l}^{t}, W_{0}\right)
\end{equation}

Here, $\Delta M_{A_{11}}^{t}$ and $\Delta D_{A_{11}}^{t}$ represent the magnitude vector difference and directional vector  difference between each task and all tasks. $t$ denotes the training rounds, and $k$ denotes the number of layers in the large language model.

\textbf{Observation1: }\textit{The directional variation of the A matrix is approximately 1.7 times greater than that of the B matrix.}

According to the findings of Tian et al.~\cite{tian2024hydralora} in their research on parameter decomposition and knowledge representation of large-scale pre-trained models, the A matrix primarily carries cross-task shared knowledge in the model parameter space, which can be regarded as the ``base framework" of global knowledge. Experiments comparing the directional vector variations between different downstream tasks and global tasks show that the amplitude of directional vector changes in the A matrix is significantly larger than that in the B matrix. This result further confirms that the directional vectors of the A matrix store more global task-related knowledge and have a direct impact on the training performance of global tasks.

\textbf{Observation2: }\textit{The magnitude variation of the B matrix is about 41 times larger than that of the A matrix.}

Concurrently, Tian et al.'s study reveals that the B matrix focuses on encoding task-specific personalized information for downstream tasks~\cite{tian2024hydralora}. In experiments comparing the magnitude vector variations between different downstream tasks and global tasks, the amplitude of magnitude vector changes in the B matrix is notably greater than that in the A matrix. This phenomenon fully demonstrates that the magnitude vectors of the B matrix store more downstream task-related knowledge and play a critical role in influencing the training effectiveness of downstream tasks.

These findings pave a clear path for optimizing model training strategies. To build globally generalizable models, training should prioritize the directional vector components of the A matrix, which enhances the model's generalization performance on global tasks. Conversely, for developing personalized models, the optimization focus should shift to the magnitude vector components of the B matrix, thereby strengthening the model's ability to encode task-specific features and driving performance breakthroughs in niche scenarios.

\begin{figure*}[!htbp]
  \centering
  \includegraphics[width=\textwidth]{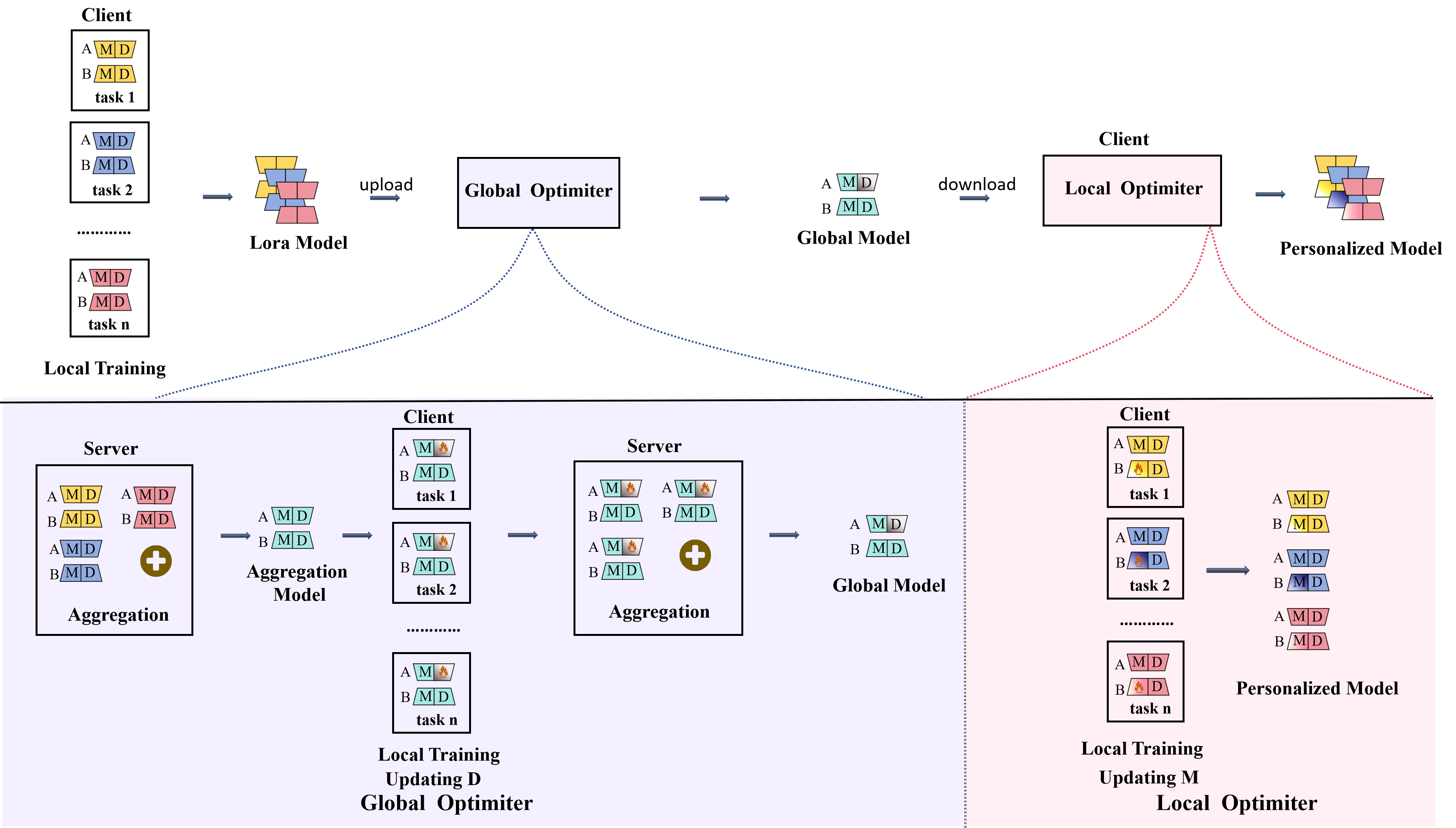}
  \caption{Architecture diagram of federated LoRA Fine-Tuning for global and local.}
  \label{fig:your_label2}
\end{figure*}

\section{Proposed Methodology}

\subsection{Outline}

Based on the observations in Chapter 2, we propose the architecture of FedLoRA-Optimizer to improve the performance of global and personalized models under data heterogeneity. This architecture addresses various fine-tuning task types by focusing on the changes in the direction and magnitude vectors of the LoRA matrices. It proposes tailored strategies to train models that meet diverse requirements in heterogeneous task settings. As shown in Figure 2, the architecture consists of a global optimizer and local optimizers. Clients upload their models to the global optimizer for global model training, and local optimizers then fine-tune the trained global model into personalized models for each task.
Next, we will elaborate on the working principles of the global optimizer and the local optimizers in detail.
\subsection{Fine-Tuning Techniques for Global Model Optimization}
In federated efficient fine-tuning under task-heterogeneous scenarios, we focus on the adjustment of directional vectors to enhance the generalization capability of the global model. Specifically, after decomposing matrices A and B into the product of directional vectors and dimensional vectors.As defined in Formula (\ref{6}), which clarifies the decomposition form of the matrices. 

\begin{equation}
    \mathbf{A} = \mathbf{A}_M \cdot \mathbf{A}_D, \quad \mathbf{B} = \mathbf{B}_M \cdot \mathbf{B}_D 
    \label{6}
\end{equation}

Then we employ the federated averaging aggregation method to aggregate the LoRA-fine-tuned matrices, preserving the shared knowledge across clients. The aggregation formulas are as shown in (\ref{7}), (\ref{8}), (\ref{9}) and (\ref{10}).

\begin{equation}
        \bar{\mathbf{A}}_D = \frac{1}{N} \sum_{i=1}^{N} \mathbf{A}_{D,i}
        \label{7}
\end{equation}
\begin{equation}
       \bar{\mathbf{A}}_M = \frac{1}{N} \sum_{i=1}^{N} \mathbf{A}_{M,i}
        \label{8}
\end{equation}
\begin{equation}
    \bar{\mathbf{B}}_M = \frac{1}{N} \sum_{i=1}^{N} \mathbf{B}_{M,i}
    \label{9}
\end{equation}
\begin{equation}
    \bar{\mathbf{B}}_D = \frac{1}{N} \sum_{i=1}^{N} \mathbf{B}_{D,i}
    \label{10}
\end{equation}

Here $\bar{\mathbf{A}}_D$ , $\bar{\mathbf{A}}_M$ , $\bar{\mathbf{B}}_D$ and $\bar{\mathbf{B}}_M$ represent the respective matrices after federal aggregation, $\mathbf{A}_M$ and $\mathbf{B}_M$ represent the magnitude vectors of matrices A and B, respectively, while $\mathbf{A}_D$ and $\mathbf{B}_D$ represent the direction vectors of matrices A and B.
$\mathbf{A}_{D,i}$ denotes the direction vector of the A matrix for the i-th client, and $\mathbf{A}_{M,i}$ denotes the magnitude component of the A matrix for the i-th client.
Similarly, $\mathbf{B}_{D,i}$ denotes the direction vector of the B matrix for the i-th client, and $\mathbf{B}_{M,i}$ denotes the magnitude component of the B matrix for the i-th client, N denotes the number of clients.

Specifically, each client fine-tunes the model using LoRA on its local data during the local training phase. After local training, during the aggregation phase, we use standard Federated Averaging to efficiently aggregate the matrices obtained from LoRA fine‑tuning. Based on this framework, the update of the global optimizer can be expressed as:
\begin{equation}
    \mathbf{W}_\text{g} = \mathbf{W}_0 + \bar{\mathbf{B}}_M \cdot \bar{\mathbf{B}}_D \cdot \bar{\mathbf{A}}_M \cdot (\bar{\mathbf{A}}_D+\Delta\mathbf{A}_{D,\text{g}}) 
\end{equation}

Here, \(\mathbf{W}_\text{g}\) represents the global model, and \( W_0 \) denotes the weights of the original model, $\Delta\mathbf{A}_{D,\text{g}}$ represents the separate adjustment of the direction of matrix A.

\subsection{Efficient Fine-Tuning Techniques for Global and Local Fusion}
To fully leverage the knowledge of the global model and develop highly personalized models, we propose introducing a local optimizer in series within the global model training pipeline. The global optimizer first completes the training of the global model, and based on this, the local optimizer conducts training for diverse personalized tasks to output task-specific models, achieving synergistic enhancement. Specifically, the global model input to the local optimizer undergoes minimal LoRA fine-tuning to adapt into a basic personalized model rapidly. Thereafter, we focus on training the magnitude module of the B matrix for personalized optimization, enabling precise matching to task-specific features. This approach enhances model performance in specific application scenarios and aims to develop more effective personalized models. The formulas for the models trained on different tasks are as follows:
\begin{equation}
\mathbf{W}_\text{l} = \mathbf{W}_\text{g} + 
    (\bar{\mathbf{B}'}_M + \Delta\mathbf{B}'_{M,\text{l}}) \cdot \bar{\mathbf{B}'}_D \cdot \bar{\mathbf{A}'}_M \cdot \bar{\mathbf{A}'}_D
\end{equation}

Here, $\mathbf{W}_\text{l}$ denotes the optimized parameters of the local model. $\mathbf{A}_M$ and $\mathbf{A}_D$, as well as $\mathbf{B}_M$ and $\mathbf{B}_D$, correspond to the magnitude and direction components of the A and B matrices in the local personalized model, respectively. $\Delta\mathbf{B}'_{M,\text{l}}$ represents the personalized adjustment applied to the magnitude component of the B matrix.

To implement the fine-tuning strategy of ``fixing the matrix \(\mathbf{A}\) and only optimizing the magnitude of matrix \(\mathbf{B}\)", we define a local loss function that balances task adaptation and parameter regularization:

\begin{equation}
\mathcal{L} \text {local}=\mathcal{L} \operatorname{task}\left(\mathbf{W}_\text{l} \mathbf{x}, \mathbf{y}\right)+\frac{\lambda}{2}\left\|\Delta \mathbf{M}_{\text {l }}\right\|_{F}^{2} \tag{11}
\end{equation}
\(\mathcal{L}_\text{local}\) denotes the local loss function, which aims to optimize the magnitude vector variation. $\mathcal{L}_{\text{task}}\left(\mathbf{W}_\text{l} \mathbf{x}, \mathbf{y}\right)$ denotes a function that measures the discrepancy between the model's prediction $\mathbf{W}_\text{l} \mathbf{x}$ and the ground-truth label $\mathbf{y}$.\(\Delta\mathbf{M}_\text{l}\) denotes the local magnitude update of matrix \(\mathbf{B}\), which is the only trainable parameter, reflecting the design of magnitude-only optimization. The parameter \(\lambda\)\ represents the regularization coefficient that suppresses overfitting of the magnitudes. Given the adjusted model parameters and input data \(\boldsymbol{\mathrm{x}}\), it generates predictive outputs, essentially embodying how the model processes inputs to produce results.\(\Vert \cdot \Vert_F^2\) represents the squared Frobenius norm of a matrix. It measures the size of the matrix \(\Delta\boldsymbol{\mathrm{M}}_{\text{l}}\). It is used in the regularization term to constrain the extent of updates to the magnitude of matrix \(\boldsymbol{\mathrm{B}}\), preventing over - aggressive adjustments that might harm the model's performance.

To optimize $\Delta\mathbf{M}_\text{local}$ during local fine-tuning, we need to compute the gradient of $\mathcal{L}_\text{local}$ with respect to $\Delta\mathbf{M}_\text{local}$. Using the chain rule to differentiate the task loss term $\mathcal{L}_\text{task}$ and the regularization term separately, we derive the gradient update formula:
\begin{equation}
\nabla_{\Delta\mathbf{M}_\text{local}} \mathcal{L}_\text{local} = \bar{\mathbf{B}'}_D \cdot \bar{\mathbf{A}'}_M \cdot \bar{\mathbf{A}'}_D \cdot \nabla_{\boldsymbol{\mathrm{y}}^\text{pred}} \mathcal{L}_\text{task} + \lambda \cdot \Delta\mathbf{M}_\text{local} \tag{12}
\end{equation}

\(\boldsymbol{\nabla_{\Delta\mathbf{M}_\text{local}} \mathcal{L}_\text{local}}\) denotes the gradient of the local loss function \(\mathcal{L}_\text{local}\) with respect to the local magnitude update \(\Delta\mathbf{M}_\text{local}\), \(\boldsymbol{\nabla_{\mathbf{y}^\text{pred}} \mathcal{L}_\text{task}}\) represents the gradient of the task loss \(\mathcal{L}_\text{task}\) with respect to the model’s predictive output \(\mathbf{y}^\text{pred}\).

\section{Experiment}
In this section, we provide a detailed description of the main experimental setup. We conducted a series of experiments on the LLMs (LLaMA2-7B and DeepSeek-7B) to evaluate the accuracy of our approach. All algorithms were implemented in Python and executed on an A800 Linux server with 100 GB of RAM and a 14-core Intel(R) Xeon(R) Gold 6348 CPU @2.60 GHz.
\subsection{Experimental Settings}
Datasets
We employed the Databricks-Dolly-15k and Natural Instructions datasets, which encompass a diverse collection of downstream tasks. To simulate a heterogeneous task environment, we selected three representative task types: language modeling, summarization, and text generation. Following standard practice, we split each dataset into 80\% training and 20\% testing subsets for model development and evaluation.

Baseline Methods
·LoRA : the standard LoRA algorithm, training only the adapter parameters while keeping all other model weights frozen.
·Prompt Tuning and Adapt Tuning\cite{matena2022merging}: two alternative lightweight fine-tuning techniques.
We compared these methods against our FedLoRA-Optimizer framework by measuring their accuracy on the three downstream tasks and the combined “all-tasks” setting. We measured answer accuracy via the semantic similarity between model outputs and target responses.

\subsection{Overall Performance}
\textbf{RQ1: How does FedLoRA-Optimizer perform compared to other fine-tuning methods under heterogeneity data?}

\begin{table*}[htbp]
    \centering
    \caption{Performance of FedLoRA - Optimizer compared to baseline methods on two datasets}
    \label{tab:combined}
    \begin{subtable}[b]{0.48\textwidth}
        \centering
        \caption{Natural Instructions Dataset}
        \captionsetup{font=small} 
        \caption*{Experimental results on physical problem - solving, question - answer, information extraction, and all task}
        \begin{tabular}{clcccc}
            \toprule
            \textbf{Model} & \textbf{Scheme} & \textbf{PH} & \textbf{QA} & \textbf{IE} & \textbf{ALL} \\
            \midrule
            \multirow{5}{*}{LLaMA2 - 7B}& LLaMA2 - 7B & 10.23 & 37.75 & 4.77 & 16.59 \\
            & Prompt - Turning & 10.76 & 46.78 & 20.76 & 17.04 \\
            & Adapt - Turning & \textbf{12.15} & 40.07 & 7.69 & 17.56 \\
            & LoRA & 11.46 & 61.69 & \textbf{22.85} & \textbf{33.04} \\
            & Our & 11.62 & \textbf{66.69} & 21.18 & 32.44 \\
            \midrule
            \multirow{5}{*}{DeepSeek - 7B} & DeepSeek & 6.16 & 4.13 & 6.08 & 5.46 \\
            & Prompt - turning & 6.23 & 4.69 & 6.31 & 5.76 \\
            & Adapt - Turning & \textbf{8.00} & 5.82 & 7.38 & 6.52 \\
            & LoRA & 6.62 & 5.74 & \textbf{10.92} & 6.00 \\
            & Our & 6.85 & \textbf{5.89} & 10.69 & \textbf{6.44} \\
            \bottomrule
        \end{tabular}
        \label{tab:sub_a}
    \end{subtable}
    \hfill
    \begin{subtable}[b]{0.48\textwidth}
        \centering
        \caption{Databricks-Dolly-15k Dataset}
        \captionsetup{font=small} 
        \caption*{Experimental results on causal reasoning, question - answer, information extraction, and all task}
        \begin{tabular}{clcccc}
            \toprule
            \textbf{Model} & \textbf{Scheme} & \textbf{Causal} & \textbf{QA} & \textbf{IE} & \textbf{ALL} \\
            \midrule
            \multirow{5}{*}{LLaMA2 - 7B} & LLaMA2 - 7B & 13.13 & 25.49 & 17.28 & 16.80 \\
            & Prompt - Turning & 15.62 & 27.76 & 18.94 & 22.95 \\
            & Adapt - Turning & 14.92 & \textbf{46.59} & 27.14 & 23.62 \\
            & LoRA & 18.59 & 40.48 & 25.91 & 25.70 \\
            & Our & \textbf{18.99} & 40.57 & \textbf{27.91} & \textbf{26.20} \\
            \midrule
            \multirow{5}{*}{DeepSeek - 7B} & DeepSeek & 2.22 & 9.62 & 6.96 & 5.00 \\
            & Prompt - turning & 3.86 & 13.63 & 8.21 & 7.85 \\
            & Adapt - Turning & 10.10 & 31.41 & 7.31 & 15.70 \\
            & LoRA & 13.54 & 30.88 & \textbf{11.30} & 18.90 \\
            & Our & \textbf{14.55} & \textbf{31.96} & 10.15 & \textbf{20.10} \\
            \bottomrule
        \end{tabular}
        \label{tab:sub_b}
    \end{subtable}
\end{table*}

Setup: For FedLoRA-Optimizer vs LoRA, we set the low-rank adapter’s rank to 8, scaling factor to 32, and lora dropout to 0.1 applied only to the query (Q) and value (V) sublayers of self-attention for a fair comparison. Prompt-Tuning and Adapt-Tuning did not utilize these LoRA-specific hyperparameters since there is no low-rank adaptation technique involved.

\textbf{LLaMA2-7B results}: In the tasks across both the Natural Instructions and Databricks-Dolly-15k datasets, we compared FedLoRA-Optimizer against baseline methods. As shown in Table \ref{tab:combined}, FedLoRA-Optimizer demonstrated superior overall accuracy compared to LoRA: on the Natural Instructions dataset, it achieved accuracies of 11.62\%, 66.69\%, 21.18\%, and 32.44\% in PH, QA, IE, and ALL tasks (vs. LoRA's 11.46\%, 61.69\%, 22.85\%, 33.04\%), with an overall accuracy improvement of 0.73\%; on the Databricks-Dolly-15k dataset, it achieved 18.99\%, 40.57\%, 27.91\%, and 26.20\% in Causal, QA, IE, and ALL tasks (vs. LoRA's 18.59\%, 40.48\%, 25.91\%, 25.70\%), showing an overall improvement of 0.75\%.

\textbf{DeepSeek-7B results}: In the tasks across both the Natural Instructions and Databricks-Dolly-15k datasets, we compared FedLoRA-Optimizer against baseline methods. As shown in Table \ref{tab:combined}, FedLoRA-Optimizer demonstrated superior overall accuracy compared to LoRA: on the Natural Instructions dataset, it achieved accuracies of 6.85\%, 5.89\%, 10.69\%, and 6.30\% (vs. LoRA's 6.62\%, 5.74\%, 10.92\%, 6.00\%), with an overall improvement of 1.11\%; on the Databricks-Dolly-15k dataset, it achieved 14.55\%, 31.96\%, 10.15\%, and 20.10\% (vs. LoRA's 13.54\%, 30.88\%, 11.30\%, 18.90\%), showing an overall improvement of 0.53\%.

\subsection{Parameter Analysis}
\textbf{RQ2: What should be the appropriate setting for the number of LoRA parameters and the rank of the matrix during training?}

In large language model fine-tuning, the hyperparameters of the LoRA technique have a significant impact on the performance of fine-tuned pre-trained models. To explore the optimal hyperparameters for the LLaMA2-7B model under the LoRA technique, we conducted instruction-tuning experiments using the Databricks-Dolly-15k Dataset. We strictly controlled the training conditions and evaluated the models using key metrics. The experimental results (Table  \ref{tab:commands}) show that models with different ranks r and configurations perform differently. After a comprehensive consideration of various metrics, the model achieves the best performance when r = 8 and n = 2. Therefore, subsequent experiments will be based on this configuration, facilitating the LLaMA2-7B model to unleash its potential in more application scenarios.

\begin{table}[t]
  \caption{Performance of instruction tuning. We used the Databricks-Dolly-15k dataset and evaluated it under different rank settings. n denotes the number of LoRAs, and r represents the rank of each LoRA.}
  \label{tab:commands}
  \centering
  \begin{tabular}{ccc}
    \toprule
    r×n & Causal & \%Parameter \\
    \midrule
    \texttt{4×1} & 14.95 & 0.0105 \\
    \texttt{8×1} & 17.37 & 0.0211 \\
    \texttt{16×1} & 17.58 & 0.0421 \\
    \texttt{8×2} & \textbf{18.59} & 0.0331 \\
    \texttt{4×4} & 17.67 & 0.0331 \\
    \bottomrule
  \end{tabular}
\end{table}

\begin{figure}[t]
  \centering
  \includegraphics[width=0.51\textwidth]{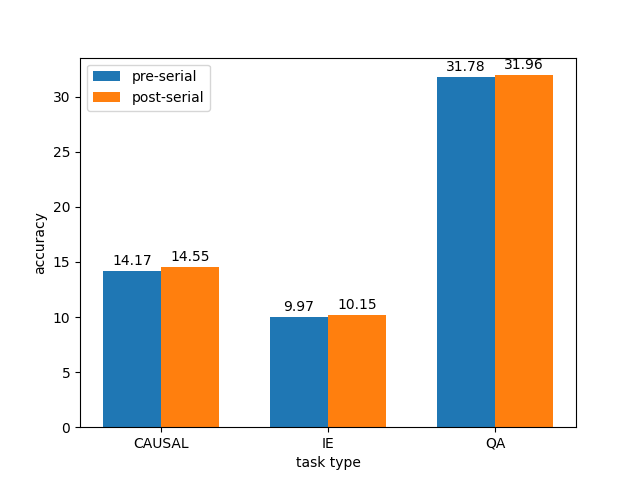}
  \caption{The comparison of accuracy across various tasks between the model that combines a global optimizer and a local optimizer in series and the model that uses only a local optimizer.In this study, we conducted experiments on the three aforementioned tasks using the DeepSeek-7B model on the Databricks-Dolly-15k dataset.}
  \label{fig:your_label3}
\end{figure}

\subsection{Ablation Studies}
\textbf{RQ3: Does the pipeline structure enhance fine-tuning?}

We compare two workflows based on the LLaMA2-7B model:\textbf{ 1. Pipeline-structured}: This model follows a sequential training strategy. It first applies a global optimizer to the LoRA-tuned model as an intermediate step. The resulting intermediate model is then passed to a local optimizer for task-specific adaptation. It forms a ``global optimization + local optimization" pipeline structure. The final personalized model is obtained by further refining the intermediate model, which can enhance task alignment and overall performance.\textbf{ 2.Non-pipeline-structured}: This model skips the global optimization stage and directly feeds the LoRA-tuned model into the local optimizer for task-specific training. Without the intermediate training step, the personalized model is adapted directly from the initial LoRA model, which may lead to suboptimal performance compared to the pipeline approach.
As shown clearly in Figure \ref{fig:your_label3}, which compares the accuracy of models trained with a combination of global and local optimizers versus those trained with only local optimizers across different task types, the models trained under the pipeline mode, represented by the orange bars labeled post-serial, consistently outperform those trained under the non-pipeline mode, represented by the blue bars labeled pre-serial, in all three tasks: causal reasoning (Causal), information extraction (IE), and question answering (QA). This result strongly supports the effectiveness of the staged training strategy.

\section{Conclusion}
We aim to design an efficient federated fine-tuning model architecture for heterogeneous environments, addressing the low fine-tuning efficiency of federated learning in such settings. This architecture integrates the classic principles of federated learning with fine-tuning methods, particularly emphasizing the control of direction and magnitude vectors. In the experimental section, we validated our approach on large language models such as LLaMA and DeepSeek, achieving the expected results successfully. Experimental outcomes indicate that our method offers particular advantages in enhancing the general(global) model’s training effectiveness and the personalized models’ performance in heterogeneous environments. However, overall gains remain limited, suggesting potential for optimization. Future research will explore optimization strategies to boost model adaptability and fine-tuning efficiency in heterogeneous settings.

\bibliographystyle{IEEEtran}
\bibliography{main} 

\end{document}